\documentclass[10pt, twocolumn, letterpaper]{article}

\usepackage{cvpr}

\usepackage{times}
\usepackage{epsfig}
\usepackage{graphicx}
\usepackage{amsmath}
\usepackage{amssymb}
\usepackage{multirow}
\usepackage[ruled]{algorithm2e}
\usepackage{dsfont}
\usepackage{epsfig}
\usepackage{subfigure}
\usepackage{url}
\usepackage{makecell}

\renewcommand{\Re}{\mathbb{R}}

 \cvprfinalcopy 

\def\cvprPaperID{7} 

\begin{document}

\title{FAST: A Framework to Accelerate Super-Resolution Processing on\\ Compressed Videos} 
%
%

\author
{
    Zhengdong Zhang, Vivienne Sze\\
    Massachusetts Institute of Technology\\
    {\tt\small \{zhangzd, sze\}@mit.edu}
}

\maketitle

\begin{abstract}
State-of-the-art super-resolution (SR) algorithms require significant computational resources to achieve real-time throughput (\eg, 60Mpixels/s for HD video). This paper introduces FAST (Free Adaptive Super-resolution via Transfer), a framework to accelerate any SR algorithm applied to compressed videos. FAST exploits the temporal correlation between adjacent frames such that SR is only applied to a subset of frames; SR pixels are then transferred to the other frames. The transferring process has negligible computation cost as it uses information already embedded in the compressed video (\eg, motion vectors and residual). Adaptive processing is used to retain accuracy when the temporal correlation is not present (\eg, occlusions). FAST accelerates state-of-the-art SR algorithms by up to 15$\times$ with a visual quality loss of 0.2dB. FAST is an important step towards real-time SR algorithms for ultra-HD displays and energy constrained devices (\eg, phones and tablets).
\end{abstract}
\section{Introduction}
\label{sec:intro}

Video content can often have lower resolution than the display, either due to the fact that there is limited communication bandwidth, or that the resolution of displays rises faster than the resolution at which video content is captured.  Thus, upsampling needs to be performed to match the resolution of the content to the display.

Today, most televisions perform upsampling using simple interpolation plus a sharpening filter~\cite{li2014comparing}; these single frame upsamplers, such as bicubic, sinc, Lanczos, Catmull-Rom~\cite{Catmull1974improving}, Mitchell-Netravali~\cite{Mitchell1988reconstruction} are based on simple splines to enable real-time throughput.  \emph{Super-resolution (SR)} can deliver higher visual quality than this by exploiting the non-local similarity of patches or by learning the mapping from the low-resolution to high-resolution from external datasets~\cite{yang2010image_history}. However, SR algorithms are computationally more expensive and slower than simple filtering. For instance, state-of-the-art neural network based SR algorithms (\eg, SRCNN~\cite{dong2014learning}) require powerful GPUs that consume hundreds of watts to achieve real-time performance~\cite{Dong_2016_ECCV,Shi_2016_CVPR}. As a result, state-of-the-art SR algorithms are typically not used in televisions and portable screens as their high computational complexity limits their throughput and results in high power consumption (\eg, the power consumption of portable devices is limited to a few watts).

\begin{figure*}
\centering{	
    \subfigure[Pipeline of FAST]{
		\includegraphics[height=0.32\textwidth]{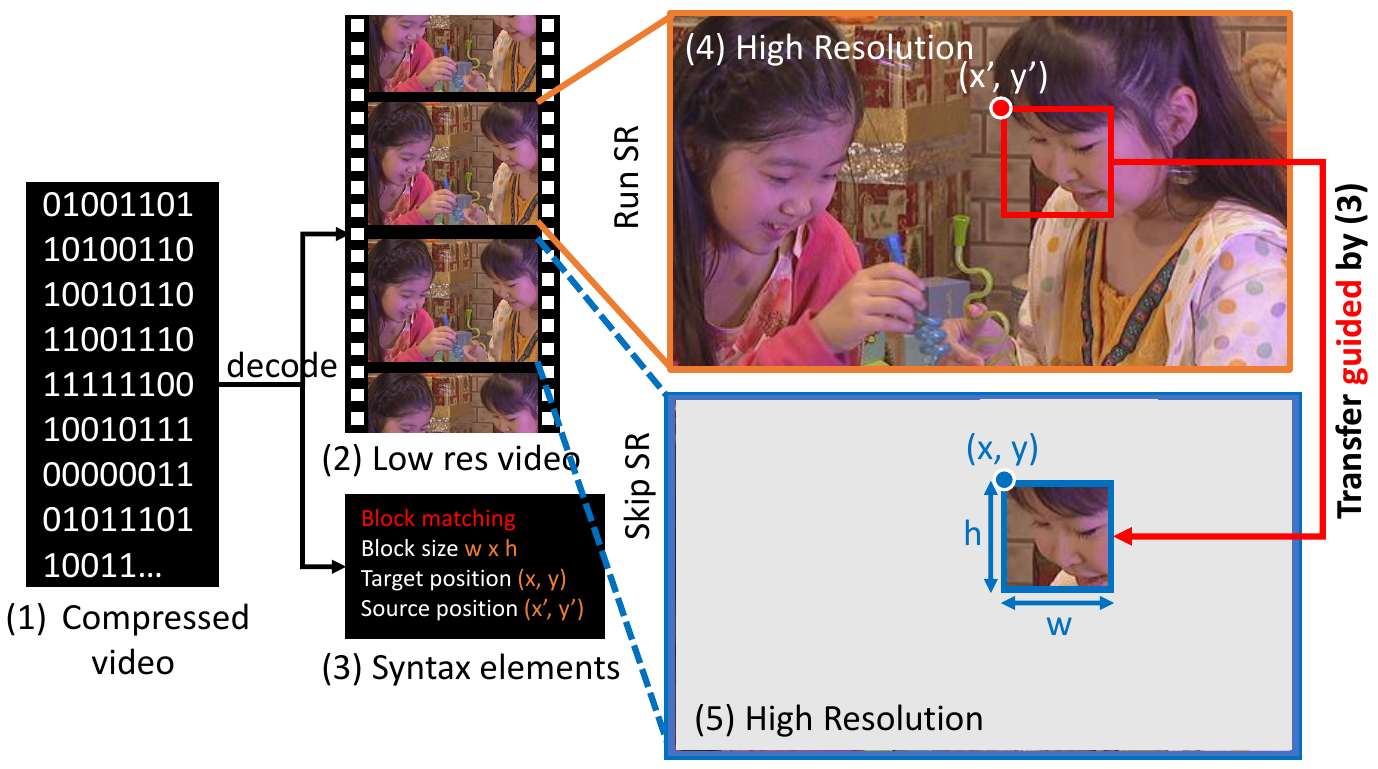}
		\label{fig:pipeline}
	}
    \subfigure[Comparison of visual quality]{
		\includegraphics[height=0.32\textwidth]{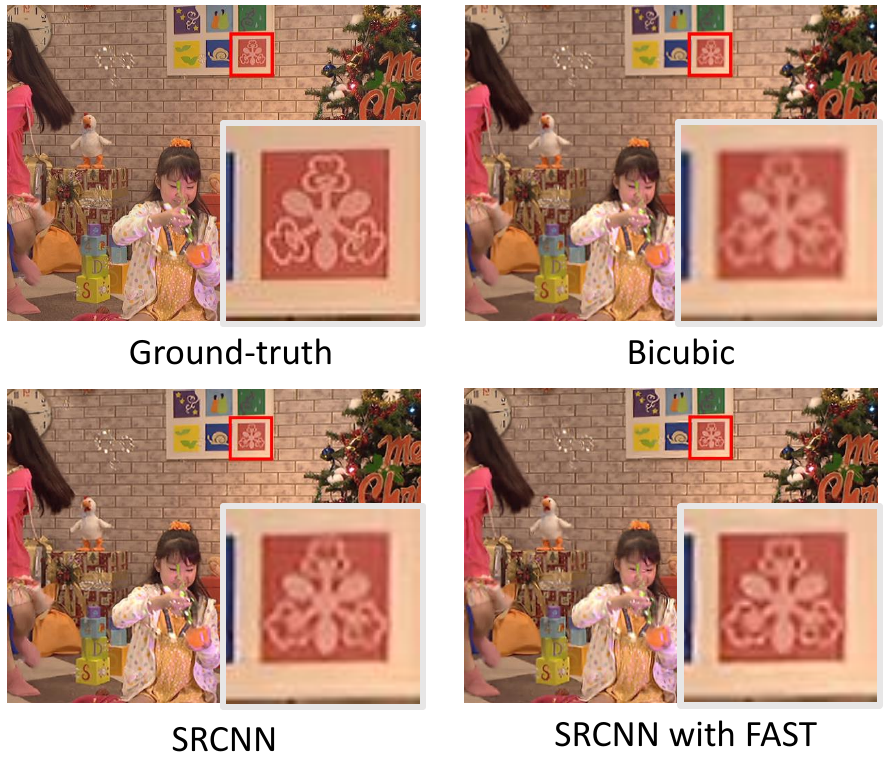}
		\label{fig:visual_quality}
	}
}
\caption{(a) \textbf{SR with FAST}: From the compressed video (1), the video decoder decompresses the low-resolution video frames (2) and the syntax elements (3). The SR algorithm generates the high resolution output for the first frame (4). Guided by the syntax elements (3), FAST adaptively transfers blocks from the first frame (4) to the second frame (5) rather than applying super-resolution on the second frame. (b) \textbf{Visual quality results}: Running SRCNN with FAST preserves the rich high frequency details that SRCNN generates compared with the blurry output of bicubic interpolation. Note that the complexity of FAST is comparable to bicubic interpolation on most of the transferred frames.}
\label{fig:teaser}
\end{figure*}

To bridge this gap, this paper proposes a technique called \emph{Free Adaptive Super-resolution via Transfer (FAST)} to accelerate existing SR algorithms.  Our approach is inspired by many concepts from the area of video coding research.  In particular, numerous techniques have been developed in video coding research to identify correlations between pixels in videos.  As a result, a compressed video is comprised of a compact description of the structure and correlation of the pixels in the video.  

The main contribution of this paper is a framework that can leverage this structure information that is available for free in the compressed video to accelerate image processing, specifically super-resolution. FAST uses this free precomputed information in two different ways:
\begin{enumerate}
\item It uses the motion vectors to exploit the inter-frame similarities between adjacent frames in a video sequence.  As a result, it can transfer the SR result from one frame to the other such that SR is only applied to a subset of frames. Since the compressed video already provides the correlation information for free, the transfer step has negligible run time and computation cost compared with the original SR algorithms. Figure~\ref{fig:pipeline} illustrates the main steps of FAST. It should be noted that the transfer step can introduce visual degradation, as motion compensation in video coding is not perfect. In this work, we propose several visual quality metrics that are easy to compute, and then adaptively enable and disable the transfer on a block-by-block basis based on these metrics.

\item FAST uses the non-overlapping block structure embedded in the compressed video, rather than the overlapping blocks traditionally used for SR; using non-overlapping blocks accelerates the processing relative to the overlapping blocks. One of the drawbacks from non-overlapping blocks is that it can result in block artifacts. To address this, an adaptive deblocking filter is applied at the block edges to remove the block artifacts while retaining the true edges in the video.

\end{enumerate}

We demonstrate the FAST framework acceleration using three well-known SR algorithms (KRR~\cite{kim2010single}, ANR~\cite{timofte2013anchored}, SRCNN~\cite{dong2014learning}) on the 20 video sequences that were used in the development of the latest video compression standard H.265/HEVC~\cite{jctvc2012test}; these sequences cover a wide range of videos including natural and screen content. On these sequences, FAST delivers up to 15$\times$ acceleration with around 0.2dB drop in PSNR.  Figure~\ref{fig:visual_quality} shows how FAST maintains the visual quality of the super-resolution using SRCNN, which is significantly better than the bicubic upsampled result, even though the complexity of FAST is comparable to bicubic upsampling on most of the transferred frames. A demo video is available at~\cite{website}.


\section{Related Work on Super Resolution}
\label{sec:previous_work}
There are two forms of super-resolution (SR) algorithms: single frame and multiple frame. Single-frame based SR algorithms are applied to each video frame independently, while multi-frame based SR algorithms are applied to  multiple frames in a small time window. FAST can be applied to either SR approach.


Sophisticated single-frame based SR algorithms often leverage machine learning techniques~\cite{freeman2000learning,freeman2002example}. Among them are sparse-representation~\cite{yang2010image_sp,zeyde2010single}, Kernel Ridge Regression (KRR)~\cite{kim2010single}, anchored neighbor regression (ANR)~\cite{timofte2013anchored}, and in-place example regression~\cite{yang2013fast}.  Recently, deep convolutional neural networks (CNNs) have been used to perform super-resolution (\eg, SRCNN~\cite{dong2014learning,bruna2015super}).  They achieve state-of-the-art results at the cost of high computational complexity; for instance, the three convolutional layers in SRCNN~\cite{dong2014learning} require $8032$ multiplications per pixel, which is significantly higher than simple interpolation with one filter. While CNNs can be accelerated by powerful GPUs to achieve real-time performance on high-definition videos ($1920\times1080$)~\cite{Shi_2016_CVPR,Dong_2016_ECCV}, these GPUs consume several hundred watts and are not suitable for embedding into televisions and portable devices (\eg, phones, tablets). 

There are also SR algorithms that exploit the self-similarities of blocks within each image~\cite{glasner2009super,huang2015single}. The proposed FAST framework shares many common insights with  in-place example regression~\cite{yang2013fast}, which performs local block prediction inside the same frame; the important distinction is that FAST uses predictions across frames, which has negligible cost when exploiting embedded information in the compressed videos.

Multiple-frame based super-resolution algorithms are largely based on the registration of neighboring frames~\cite{baker1999super}. Many of these algorithms are iterative, including the Bayesian based approach~\cite{liu2011bayesian}, and the $\ell_1$-regularized total variation based approach~\cite{mitzel2009video}.  At the same time, there are non-iterative methods that avoid registration with non-local mean~\cite{protter2009generalizing}, 3D steer kernel regression~\cite{takeda2009super}, and self-similarity \cite{jeong2015multi}. Deep neural networks can also be used in the form of bidirectional recurrent convolutional networks~\cite{huang2015bidirectional}, and deep draft-ensemble learning~\cite{liao2015video}. Spatial-temporal coherence of multiple-frame based SR is addressed in~\cite{hsu2015temporally} and~\cite{caballero2016real}. Similar to this work, motion compensation is also used in~\cite{caballero2016real} with two key differences: (1) both the original and motion compensated frames are fed into the neural network, while in FAST only the first frame is processed by the neural network; (2) the motion vectors are explicitly computed with optical flow, while FAST uses the motion vectors embedded in the compressed video. Thus, techniques proposed in FAST could potentially help accelerate~\cite{caballero2016real}. 

\section{Video Coding Basics} 

The FAST framework is inspired by many concepts from video coding research; in particular, the use of temporal correlation and the adaptive block-based processing.  Temporal correlation is used in video coding during a process called motion compensation. Motion compensation is widely used in popular video coding standards including MPEG-2~\cite{le1991mpeg}, used for HDTV broadcast; H.264/AVC~\cite{h264-standard}, used for most video content on the Internet; and H.265/HEVC~\cite{hevc-standard}, the most recent standard. Motion compensation involves predicting a block of pixels in the current frame from a block of pixels in a temporally-adjacent, previously-encoded frame. 

During compression, videos are divided into non-overlapping blocks and encoded on a block-by-block basis. At the video encoder, motion estimation is used to determine which block of pixels in the previously encoded frame would best match the current block for minimum prediction error. As a result, the encoder only needs to signal the small difference between the two blocks, which typically requires fewer bits than signaling the original pixels; this difference is referred to as the \emph{residual}. The offset between the location of the source block in the previous frame and the location of the target block in the current frame is called a \emph{motion vector}.  Both the residual and motion vector are part of the \emph{syntax elements} that are embedded in the compressed video to express the structure of the video as shown in Figure~\ref{fig:pipeline}. The video decoder uses these syntax elements to reconstruct the video; for instance, the motion compensation at the decoder uses the motion vector to determine which pixels to use from the previously encoded frame to generate the source block, and adds the residual to reconstruct the target block. 

Video encoders divide up a video into sets of frames called group-of-pictures (GOP); motion compensation is then applied to frames within the same GOP.  Each GOP typically contains between 6 to 16 frames that are visually similar with no scene transitions~\cite{FinalCutProGOP}.  FAST will use the GOP size to define the maximum distance for transfer. The blocks of pixels in the first frame of a GOP are encoded by only exploiting spatial correlation (\ie, correlation with neighboring pixels within the frame), which is referred to as intra-predicted blocks; these frames are referred to as I frames.  The subsequent frames are primarily composed of blocks that exploit temporal correlation, referred to as motion compensated blocks; these frames are referred to as P frames and may contain some intra-predicted blocks in cases of low temporal correlation. 




\section{FAST Framework}
\label{sec:fast_transfer}
The FAST framework consists of the following features:
\begin{itemize}
\item Transfer SR pixels using motion compensation
\item Adaptive transfer to retain visual quality (\ie, PNSR)
\item Apply SR to non-overlapping blocks and remove block artifacts with an adaptive deblocking filter 
\end{itemize}

\subsection{Transfer using Motion Compensation} 
\label{sec:fast_formulation}

To perform a transfer, the SR pixels are copied based on the scaled motion vector to the previous frame; here, the \emph{scaled} motion vector refers to the fact that the motion vector extracted from the compressed video is multiplied by 2$\times$ if FAST is performing a 2$\times$ upsampling.  If the scaled motion vector is fractional, an interpolation filter is applied to the SR pixels; FAST uses the same interpolation filter as the one used for fractional motion vectors during motion compensation for video decompression~\cite{hevc-book}. The residual is upsampled using bicubic interpolation and added to the transferred SR pixels.

Recall that the video encoder uses intra-prediction for blocks when there is not sufficient temporal correlation. In these cases, transfer using motion compensation cannot be used. However, with the exception of the first frame in the GOP (an I frame), intra-predicted blocks rarely occur in the subsequent P frames; this makes sense, as otherwise it would lead to poor coding efficiency of the video. Since there are few intra-predicted blocks, FAST upsamples them with bicubic upsampling.

In summary, FAST only applies the expensive SR processing on the first frame of a GOP (except when the transfer needs to be reset as discussed in Section~\ref{sec:transfer_reset}). For all other frames in the GOP, FAST uses lightweight processing such as interpolation and bicubic upsampling. Since a typical GOP size is on the order of 6 to 16 frames~\cite{FinalCutProGOP}, the majority of the frames are accelerated with lightweight processing; the overall acceleration depends on the GOP size.  Section~\ref{sec:results} reports the visual quality impact and acceleration of FAST for various GOP sizes and video sequences.

\subsection{Adaptive Transfer}
\label{sec:adaptive}


Transferring using motion compensation is not perfect. Error is introduced due to (1) poor temporal correlation where the motion compensated SR pixels are not good predictors; (2) the upsampling of the residual using bicubic interpolation does not account for potential high frequency components in the prediction error. The error from adaptive transfer is more noticeable when the quantization used during compression is low since there are more high frequency components in the residual that are not preserved by the bicubic interpolation.  

\begin{figure*}
\centering{	
    \subfigure[Adaptive transfer to avoid artifacts]{
		\includegraphics[height=0.3\textwidth]{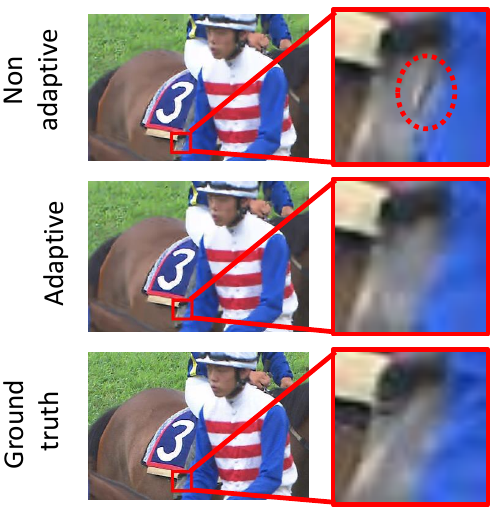}
		\label{fig:adapt_vis}
	}
    \subfigure[Cause of the artifact]{
		\includegraphics[height=0.3\textwidth]{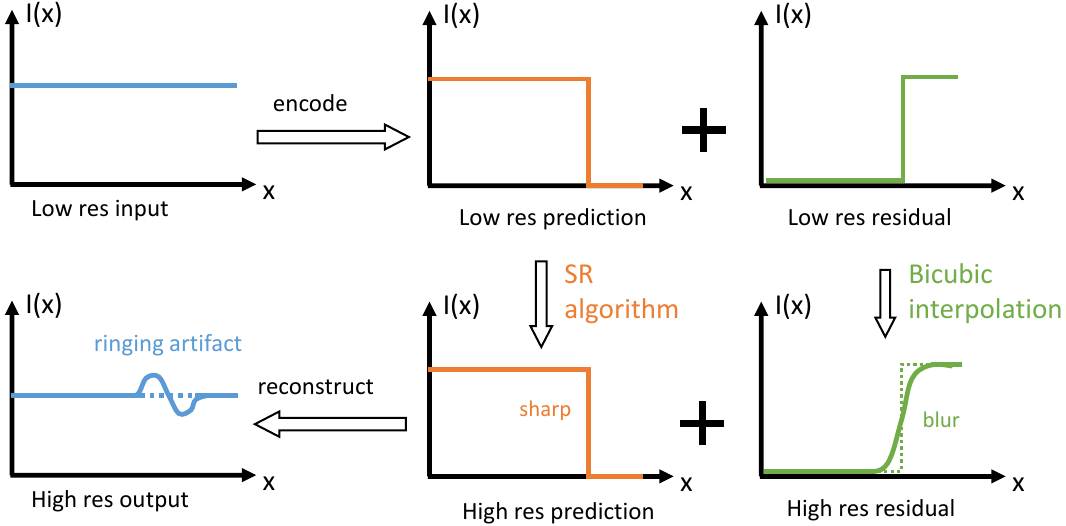}
		\label{fig:adapt_ex}
	}

}
\caption{(a) Adaptive transfer avoids the ringing artifact in flat areas of the output of non-adaptive transfer. (b) Sharp edges in the residual causes ringing artifacts.}
\label{fig:adap_illustration}
\end{figure*}

Figure~\ref{fig:adapt_vis} shows an example of when the error from the transfer, specifically the bicubic upsampling of the residual, can cause ringing artifacts. These types of errors occur when the source block has a sharp edge and the target block should be smooth; this mismatch between source and target can occur due to the rate-distortion optimization in the video encoder~\cite{sullivan1998rate}, where both the visual quality and the bits of the syntax elements must be considered (\eg, trade-off fewer bits for the motion vector for more distortion and larger residual). The bicubic upsampling smooths out the sharp edge in the residual, which results in an incorrect sharp edge in reconstructed output as shown in Figure~\ref{fig:adapt_ex}.  

\subsubsection{Deciding which blocks to transfer}
In order to maintain visual quality, only a subset of the blocks in a frame are transferred. These blocks are adaptively selected based on the magnitude of the residual; note the value of the residual can be obtained for free from the video decoder.  Specifically, transfer is disabled for blocks with large residuals (\ie, \emph{mean absolute value of their residual}); these disabled blocks are upsampled with bicubic interpolation. A threshold $\eta$ is used to detect this blocks; the value of $\eta$ is trained on the Middlebury stereo dataset~\cite{Middlebury} to maximized the PSNR across all blocks. In all the experiments presented in this paper, we set $\eta = 10$. Figure~\ref{fig:adapt_vis} shows how this simple threshold avoids the ringing artifact in the FAST result; this approach increases the PSNR by around 0.17dB.  Unlike other approaches that use thresholding for acceleration~\cite{yang2012coupled,yang2013fast}, FAST uses thresholding to improve visual quality.



\subsubsection{Deciding when to reset transfer}
\label{sec:transfer_reset}
While the residual error may be low for a given frame, it can accumulate as the number of transferred frames increase. This is due to the fact that the frames can be chained in a I-P-P-P GOP structure. Thus, the transfer should also be adaptively reset when the accumulated error gets too large.  This accumulated error needs to be carefully modeled as it involves the interaction of error in different frames. As discussed earlier, the upsampling of the residual using bicubic interpolation does not account for potential high frequency components in the prediction error; accordingly, we use the Laplacian of the residual which has a higher correlation (0.484) with the PSNR drop compared with other metrics such as the magnitude of the residual (0.321).  Thus the accumulated high frequency error can be modeled as the accumulated error of its source block from the previous frame plus the Laplacian of the residual for the target block in the current frame. This computation only involves simple filtering with a Laplacian filter and an addition to accumulate the error.  When this error exceeds a given threshold, the transfer is reset, and the accumulated error is reset to zero.   

The reset involves applying super-resolution on the low resolution decoded block and thus has an impact on the overall throughput. Sweeping the reset threshold gives an accuracy versus speed trade-off which is discussed in Section~\ref{sec:results} and shown in Figure~\ref{fig:qp_22_threshold}.

\subsection{Non-Overlapping Blocks}
\label{sec:block}
%

\begin{figure}
\centering{	
    \subfigure[Block structure]{
		\includegraphics[height=0.3\textwidth]{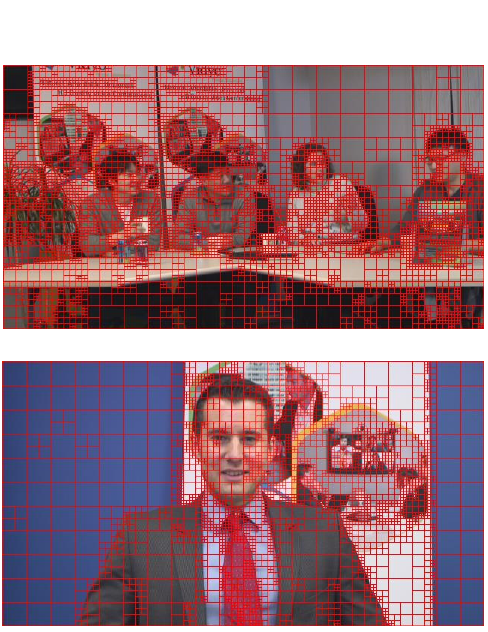}
		\label{fig:block_structure}
	}
    \subfigure[Examples of deblocking]{
		\includegraphics[height=0.3\textwidth]{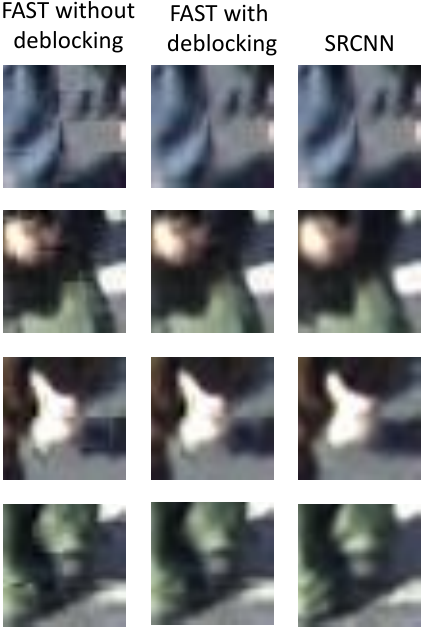}
		\label{fig:deblock_vis}
	}

}
\caption{(a) An image is adaptively divided into non-overlapping blocks, with larger blocks corresponding to simple and well-predicted content. (b) Examples of running SRCNN with FAST before and after deblocking, compared to SRCNN results on the second frame. \textit{(Best viewed in color)}}
\label{fig:deblocking}
\end{figure}

Most existing SR algorithms divide an image into densely overlapped blocks, and average the outputs of these overlapped blocks to avoid discontinuities at the block boundaries. This is computationally expensive since each pixel in a frame is processed multiple times since it belongs to multiple blocks~\cite{freeman2002example,yang2010image_sp,yang2012coupled,yang2013fast}. In contrast, video coding uses non-overlapping block, as shown in Figure~\ref{fig:block_structure}, so that each pixel is covered by exactly one block. FAST uses this non-overlapping block structure, where each pixel is only processed once, which significantly reduces the computational complexity of the transferred frames.

The drawback of using non-overlapping blocks is that it can introduce artificial edges at the block boundaries. FAST addresses this by applying an adaptive deblocking filter, specifically the one used in H.265/HEVC~\cite{norkin2012hevc}, on the block edges; adaptive deblocking filters are widely used in video coding and have been shown to give similar visual quality~\cite{hevc-book} as using overlapped blocks~\cite{malvar1990lapped}. The main objective of the deblocking filter is to remove the artificial edges on the block boundary due to non-overlapping block-based coding while keeping the true edges. Adaptive processing is used to decide whether to apply the deblocking filter to each block edge, and if so, decide the smoothing strength of the filter.  This decision is based on the differences in coding mode and pixels on both sides of the block boundary including motion vectors, whether intra-prediction was used, the number of non-zero coefficients, the amount of quantization, and the variations of the pixels~\cite{norkin2012hevc}. 

Using the H.265/HEVC deblocking filter to remove the artificial edges, as shown in Figure~\ref{fig:deblock_vis}, increases the PSNR of the output by around 0.3dB, which is only 0.03dB less than using SRCNN with overlapping blocks. At the same time, using non-overlapping blocks can give over an order of magnitude speed up as compared using overlapping blocks (\eg, each pixel in the first convolutional layer of SRCNN is processed 81 times).  The deblocking filter is lightweight, and accounts for only $17\%$ of the video decoder processing time on an ARM CPU~\cite{bossen2012hevc}, and only $3\%$ of the video decoder power consumption in specialized hardware~\cite{tikekar2014249}.


\section{Experimental Results}
\label{sec:results}


\subsection{Evaluation Dataset and Setup}

We evaluate the FAST framework on the 20 video sequences used in the development of the latest video coding standard H.265/HEVC, as they cover a wide range of video content~\cite{jctvc2012test}. We use the original uncompressed videos as the high-resolution ground-truth. We then synthetically downsample them to lower-resolution, and encode them with the H.265/HEVC encoder~\cite{sullivan-tcsvt-2012}. The FAST framework uses the decompressed low resolution frames and the decoded syntax elements as the input, which can be obtained for free from the video decoder.

We evaluate the FAST framework on three SR algorithms: KRR~\cite{kim2010single}, ANR~\cite{timofte2013anchored} and SRCNN~\cite{dong2014learning}\footnote{Only the MATLAB code for SRCNN is publicly available; thus, our run times are higher than numbers reported in~\cite{dong2014learning} which used C code.}. For each algorithm, we conduct two experiments: (1) we run the SR algorithm directly on all of the low-resolution frames; (2) we use the SR algorithm to upsample the first frame, then use FAST to transfer the result to all the other frames. 

For quantitative evaluation, we compute the PSNR between each output frame and the ground-truth high-resolution frame. Since the ground-truth frames are not compressed, the reported PSNR includes both the quantization noise from lossy compression and the quality of the super-resolution algorithm. As a result, the reported gain of SR algorithms against bicubic interpolation in our experiments is around 1 dB lower than the gains reported in the original papers. We measure the run time to quantify the FAST acceleration using MATLAB on a 3.3GHz Xeon CPU.



\subsection{Evaluation Results}
\label{sec:results_QP_27}

\begin{figure*}
\includegraphics[width=\textwidth]{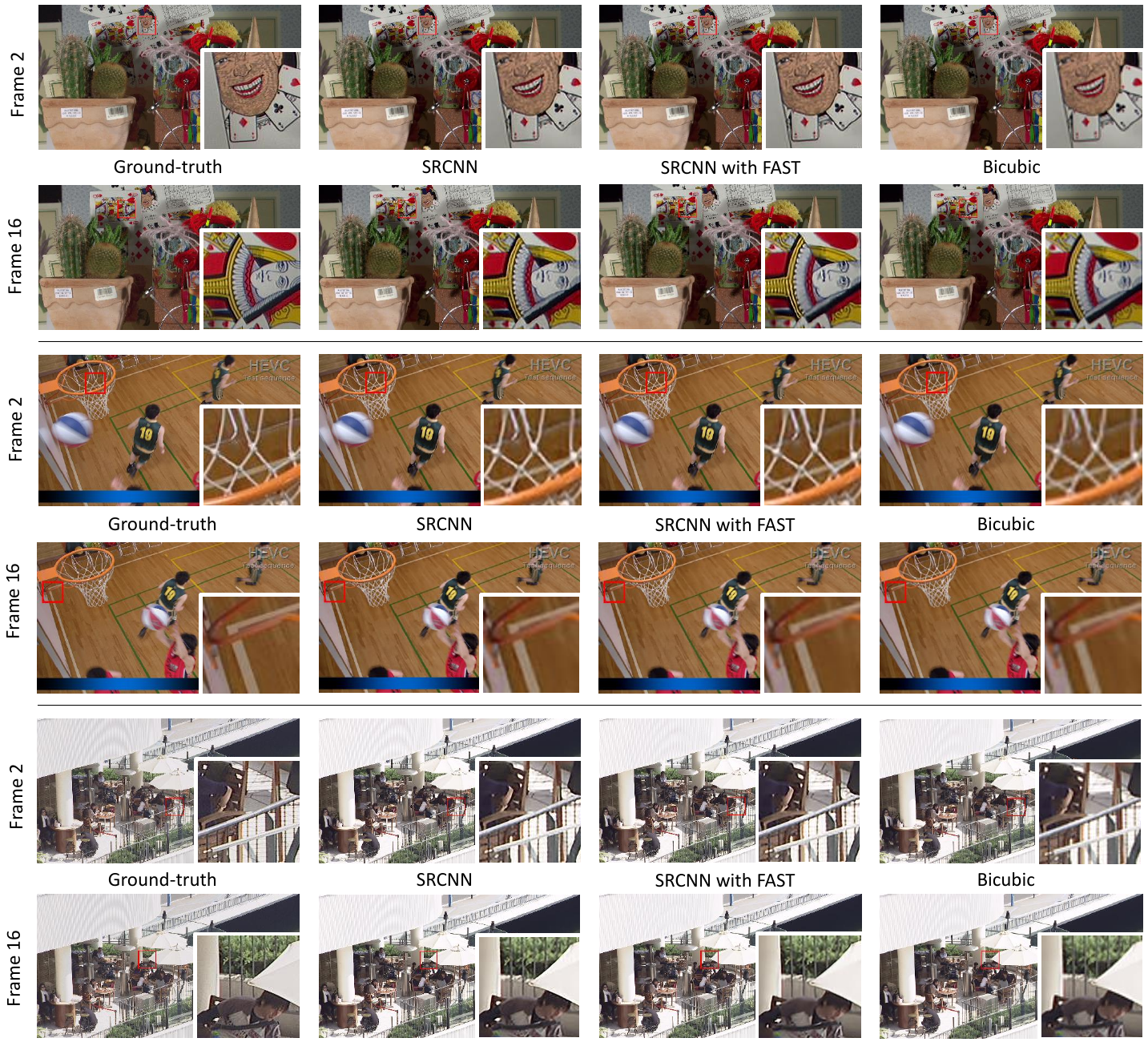}
\caption{Running different SR algorithms with FAST on different frames of different sequences. Top: \emph{Cactus}. Mid: \emph{BasketballDrill}. Bottom: \emph{BQTerrace}. Note how FAST maintains the appearance of SR output. \textit{(Best viewed in color)}}
\label{fig:vis_cmp}
\end{figure*}

Table~\ref{tab:all_4} and Table~\ref{tab:all_16} shows the PSNR and acceleration ratio for the various SR algorithms, when FAST transfers all the way to the $4^{th}$ and $16^{th}$ frame, respectively. Figure~\ref{fig:vis_cmp} shows the visual quality of the FAST framework on SRCNN for both the $2^{nd}$ and the $16^{th}$ frames. From these results we can see that the FAST framework is robust as it maintains the PSNR and achieves significant acceleration across all evaluated video sequences and SR algorithms. 

\begin{table*}
\resizebox{\textwidth}{!}{
\begin{tabular}{|c|c|c|c|c|c|c|c|c|c|c|c|}
\hline
\multirow{2}{*}[-0.5em]{\makecell{Sequence name}} & \multirow{2}{*}[-0.5em]{Size} & \multicolumn{3}{c|}{KRR} & \multicolumn{3}{c|}{SRCNN} & \multicolumn{3}{c|}{ANR} & \multirow{2}{*}[-0.7em]{\makecell{Bicubic\\PSNR}}\\\cline{3-11}
& & \makecell{SR\\PSNR} & \makecell{FAST\\PSNR} & \makecell{Speed\\up} & \makecell{SR\\PSNR} & \makecell{FAST\\PSNR} & \makecell{Speed\\up} & \makecell{SR\\PSNR} & \makecell{FAST\\PSNR} & \makecell{Speed\\up} & \\
\hline
BQMall & 416$\times$240 & 28.67 & 28.63 & 3.9$\times$ & 28.92 & 28.88 & 4.1$\times$ & 28.5 & 28.48 & 4.1$\times$ & 27.65 \\\hline
BQSquare & 208$\times$112 & 24.1 & 24.16 & 3.9$\times$ & 24.78 & 24.85 & 4.1$\times$ & 23.67 & 23.75 & 4.1$\times$ & 22.77 \\\hline
BQTerrace & 960$\times$528 & 29.01 & 29.22 & 3.9$\times$ & 29.37 & 29.51 & 4.0$\times$ & 28.73 & 28.97 & 3.9$\times$ & 27.45 \\\hline
BasketballDrill & 416$\times$240 & 31.7 & 31.69 & 3.9$\times$ & 31.6 & 31.6 & 3.6$\times$ & 31.18 & 31.18 & 3.8$\times$ & 30.19 \\\hline
BasketballDrillTe$\times$t & 416$\times$240 & 31.7 & 31.69 & 3.9$\times$ & 31.6 & 31.6 & 4.1$\times$ & 31.18 & 31.18 & 3.8$\times$ & 30.19 \\\hline
BasketballDrive & 960$\times$528 & 34.51 & 34.51 & 3.6$\times$ & 34.65 & 34.65 & 3.6$\times$ & 34.19 & 34.21 & 3.9$\times$ & 33.49 \\\hline
BasketballPass & 208$\times$112 & 30.54 & 30.58 & 3.9$\times$ & 30.63 & 30.65 & 3.7$\times$ & 30.34 & 30.4 & 3.6$\times$ & 29.7 \\\hline
BlowingBubbles & 208$\times$112 & 28.68 & 28.76 & 3.8$\times$ & 28.68 & 28.74 & 3.7$\times$ & 28.52 & 28.57 & 3.7$\times$ & 27.91 \\\hline
Cactus & 208$\times$112 & 32.68 & 32.8 & 3.9$\times$ & 32.72 & 32.83 & 3.5$\times$ & 32.59 & 32.7 & 3.9$\times$ & 31.78 \\\hline
ChinaSpeed & 960$\times$528 & 24.7 & 24.66 & 3.9$\times$ & 24.9 & 24.83 & 4.0$\times$ & 24.52 & 24.52 & 4.4$\times$ & 23.97 \\\hline
FourPeople & 512$\times$384 & 34.43 & 34.43 & 4.1$\times$ & 34.75 & 34.76 & 3.9$\times$ & 34.36 & 34.36 & 3.9$\times$ & 33 \\\hline
Johnny & 640$\times$352 & 37.08 & 37.09 & 4.0$\times$ & 37.22 & 37.23 & 3.5$\times$ & 36.81 & 36.84 & 4.1$\times$ & 35.76 \\\hline
Kimono & 960$\times$528 & 37.5 & 37.44 & 3.7$\times$ & 37.36 & 37.38 & 3.7$\times$ & 37.44 & 37.4 & 3.9$\times$ & 37.56 \\\hline
KristenAndSara & 640$\times$352 & 34.98 & 34.98 & 3.9$\times$ & 35.84 & 35.85 & 4.3$\times$ & 34.68 & 34.69 & 3.9$\times$ & 33.22 \\\hline
ParkScene & 960$\times$528 & 33.35 & 33.36 & 3.9$\times$ & 33.35 & 33.37 & 3.7$\times$ & 33.33 & 33.36 & 4.0$\times$ & 32.98 \\\hline
PartyScene & 416$\times$240 & 25.41 & 25.4 & 3.9$\times$ & 25.58 & 25.58 & 4.3$\times$ & 25.24 & 25.23 & 3.4$\times$ & 24.55 \\\hline
PeopleOnStreet & 1280$\times$800 & 33.54 & 33.34 & 3.8$\times$ & 33.35 & 33.19 & 4.0$\times$ & 33.67 & 33.45 & 3.5$\times$ & 32.89 \\\hline
RaceHorses & 416$\times$240 & 29.9 & 29.79 & 3.6$\times$ & 29.95 & 29.82 & 3.9$\times$ & 29.77 & 29.7 & 2.8$\times$ & 29.02 \\\hline
SlideEditing & 640$\times$352 & 19.93 & 19.93 & 4.0$\times$ & 20.73 & 20.73 & 3.5$\times$ & 19.86 & 19.87 & 3.6$\times$ & 19.3 \\\hline
Traffic & 1280$\times$800 & 34.73 & 34.71 & 3.9$\times$ & 34.73 & 34.72 & 4.0$\times$ & 34.75 & 34.74 & 3.9$\times$ & 33.97 \\\hline
\bf Average &  & \bf30.86 & \bf30.86 & \bf3.9$\times$ & \bf31.04 & \bf31.04 & \bf3.9$\times$ & \bf30.67 & \bf30.68 & \bf3.8$\times$ & \bf29.87 \\\hline
\end{tabular}
}
\caption{With 3 transfers from the $1^{st}$ frame to the $4^{th}$ frame, FAST gets around 4$\times$ speed up uniformly across all sequences for all SR algorithms, with no visual quality loss.}
\label{tab:all_4}
\end{table*}

\begin{table*}
\resizebox{\textwidth}{!}{
\begin{tabular}{|c|c|c|c|c|c|c|c|c|c|c|c|}
\hline
\multirow{2}{*}[-0.5em]{\makecell{Sequence name}} & \multirow{2}{*}[-0.5em]{Size} & \multicolumn{3}{c|}{KRR} & \multicolumn{3}{c|}{SRCNN} & \multicolumn{3}{c|}{ANR} & \multirow{2}{*}[-0.7em]{\makecell{Bicubic\\PSNR}}\\\cline{3-11}
& & \makecell{SR\\PSNR} & \makecell{FAST\\PSNR} & \makecell{Speed\\up} & \makecell{SR\\PSNR} & \makecell{FAST\\PSNR} & \makecell{Speed\\up} & \makecell{SR\\PSNR} & \makecell{FAST\\PSNR} & \makecell{Speed\\up} & \\
\hline
BQMall & 416$\times$240 & 28.63 & 28.4 & 14.1$\times$ & 28.85 & 28.6 & 15.2$\times$ & 28.47 & 28.28 & 13.2$\times$ & 27.68\\\hline
BQSquare & 208$\times$112 & 23.98 & 24.1 & 14.7$\times$ & 24.65 & 24.61 & 13.7$\times$ & 23.59 & 23.78 & 9.2$\times$ & 22.7\\\hline
BQTerrace & 960$\times$528 & 28.98 & 29.16 & 14.9$\times$ & 29.34 & 29.36 & 15.9$\times$ & 28.69 & 28.95 & 14.8$\times$ & 27.43\\\hline
BasketballDrill & 416$\times$240 & 31.53 & 31.28 & 13.3$\times$ & 31.41 & 31.21 & 12.7$\times$ & 31.03 & 30.87 & 11.2$\times$ & 30.08\\\hline
BasketballDrillText & 416$\times$240 & 31.53 & 31.28 & 13.3$\times$ & 31.41 & 31.21 & 13.9$\times$ & 31.03 & 30.87 & 11.5$\times$ & 30.08\\\hline
BasketballDrive & 960$\times$528 & 34.43 & 34.22 & 12.1$\times$ & 34.56 & 34.3 & 15.3$\times$ & 34.13 & 34.02 & 14.5$\times$ & 33.47\\\hline
BasketballPass & 208$\times$112 & 30.56 & 30.26 & 13.9$\times$ & 30.62 & 30.25 & 13.3$\times$ & 30.37 & 30.18 & 8.7$\times$ & 29.81\\\hline
BlowingBubbles & 208$\times$112 & 28.45 & 28.35 & 14$\times$ & 28.47 & 28.34 & 12.3$\times$ & 28.31 & 28.25 & 8$\times$ & 27.75\\\hline
Cactus & 208$\times$112 & 32.46 & 32.36 & 14.1$\times$ & 32.52 & 32.39 & 13.9$\times$ & 32.37 & 32.32 & 14.4$\times$ & 31.6\\\hline
ChinaSpeed & 960$\times$528 & 24.79 & 24.65 & 13.8$\times$ & 24.97 & 24.81 & 13.7$\times$ & 24.61 & 24.54 & 15$\times$ & 24.05\\\hline
FourPeople & 512$\times$384 & 34.33 & 34.23 & 15.1$\times$ & 34.64 & 34.55 & 15.1$\times$ & 34.26 & 34.17 & 15.2$\times$ & 32.91\\\hline
Johnny & 640$\times$352 & 36.92 & 36.76 & 15.1$\times$ & 37.06 & 36.89 & 14.8$\times$ & 36.62 & 36.51 & 16.1$\times$ & 35.54\\\hline
Kimono & 960$\times$528 & 37.08 & 36.59 & 13.5$\times$ & 36.94 & 36.56 & 13.5$\times$ & 37.02 & 36.57 & 14.2$\times$ & 37.17\\\hline
KristenAndSara & 640$\times$352 & 34.9 & 34.78 & 14$\times$ & 35.74 & 35.6 & 17.6$\times$ & 34.6 & 34.51 & 15.4$\times$ & 33.15\\\hline
ParkScene & 960$\times$528 & 33.08 & 32.83 & 13.3$\times$ & 33.07 & 32.82 & 13.4$\times$ & 33.06 & 32.84 & 14.8$\times$ & 32.74\\\hline
PartyScene & 416$\times$240 & 25.01 & 24.85 & 13.3$\times$ & 25.17 & 24.95 & 14$\times$ & 24.88 & 24.78 & 9.3$\times$ & 24.28\\\hline
PeopleOnStreet & 1280$\times$800 & 33.35 & 32.48 & 12.7$\times$ & 33.15 & 32.38 & 15.3$\times$ & 33.47 & 32.55 & 9.4$\times$ & 32.74\\\hline
RaceHorses & 416$\times$240 & 29.91 & 29.31 & 12$\times$ & 29.95 & 29.31 & 10.7$\times$ & 29.78 & 29.31 & 4.4$\times$ & 29.1\\\hline
SlideEditing & 640$\times$352 & 19.88 & 19.84 & 16$\times$ & 20.7 & 20.61 & 13.3$\times$ & 19.8 & 19.79 & 11.5$\times$ & 19.24\\\hline
Traffic & 1280$\times$800 & 34.52 & 34.25 & 14.1$\times$ & 34.51 & 34.25 & 15.8$\times$ & 34.51 & 34.27 & 12.8$\times$ & 33.78\\\hline
\bf Average &  & \bf 30.72 &\bf 30.5 &\bf 13.9$\times$ & \bf 30.89 & \bf 30.65 & \bf 14.2$\times$ & \bf 30.53 & \bf 30.37 & \bf 12.2$\times$ & \bf 29.77\\\hline

\end{tabular}
}
\caption{With 15 transfers from the $1^{st}$ frame to the $16^{th}$ frame, FAST gets more than 10$\times$ speed up on average over all sequences for all SR algorithms, with around 0.2dB PSNR loss. Nevertheless, the PSNR of FAST output is still significantly higher than the bicubic output. Note that the complexity of FAST is comparable to bicubic interpolation on most of the transferred frames.}
\label{tab:all_16}
\end{table*}

\begin{figure}
\begin{center}
\includegraphics[width=0.8\columnwidth]{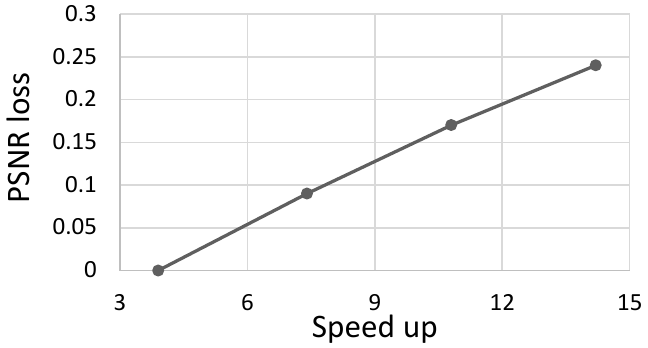}
\end{center}
\caption{The higher the acceleration ratio, the more PSNR loss, albeit the PSNR loss is only around $0.2$dB over all sequences even near $15\times$ acceleration.}
\label{fig:trade_off}
\end{figure}

Figure~\ref{fig:trade_off} shows that as the number of transferred frames increases (\ie, GOP size), the acceleration increases, but so does the drop on in PSNR.  This can be explained by looking at the breakdown in PSNR and acceleration per frame as shown in Table~\ref{tab:psnr_four_people} and Table~\ref{tab:time_four_people}, respectively. In the first four frames, FAST gives comparable PSNR as SRCNN. At the $16^{th}$ frame, FAST gives slightly worse PSNR ($<0.2$ dB) than SRCNN, which is still significantly better than bicubic even though the transferred frames in FAST have similar complexity to bicubic. The average run time on the first 4 and 16 frames is also included. We can see that the cost of the transfer is negligible compared with SRCNN. Therefore, the average run time per frame is approximately the run time of applying SR to the first frame divided by the number of processed frames.

Note that in our evaluation we are reporting the PSNR drop for the worst case, where all frames are chained together in a I-P-P-P GOP structure. For other more flexible GOP structures, such as those that allow bi-prediction, we would expect the PSNR to reduce more slowly than I-P-P-P.

\begin{table*}
{\resizebox{\textwidth}{!}{\begin{tabular}{|c|c|c|c|c|c|c|c|c|c|c|c|c|c|c|c|c|c|c|}
\hline
Frame ID & 1 & 2 & 3 & 4 & 5 & 6 & 7 & 8 & 9 & 10 & 11 & 12 & 13 & 14 & 15 & 16 & \textbf{Mean 4} & \textbf{Mean 16}\\
\hline
SRCNN & 34.97 & 34.75 & 34.71 & 34.57 & 34.86 & 34.65 & 34.6 & 34.46 & 34.78 & 34.62 & 34.56 & 34.43 & 34.74 & 34.57 & 34.51 & 34.4 & 34.75 & 34.64\\
\hline
FAST & 34.97 & 34.76 & 34.73 & 34.59 & 34.79 & 34.61 & 34.53 & 34.42 & 34.63 & 34.5 & 34.43 & 34.32 & 34.54 & 34.39 & 34.33 & 34.24 & 34.76 & 34.55\\
\hline
Bicubic & 33.1 & 33.01 & 32.98 & 32.91 & 33.01 & 32.91 & 32.87 & 32.8 & 32.97 & 32.89 & 32.84 & 32.79 & 32.95 & 32.87 & 32.82 & 32.78 & 33 & 32.91\\
\hline
\end{tabular}}}
\caption{PSNR (in dB) and visual quality loss of the FAST framework using SRCNN on the test sequence "FourPeople".}
\label{tab:psnr_four_people}
\end{table*}

\begin{table*}
{\resizebox{\textwidth}{!}{\begin{tabular}{|c|c|c|c|c|c|c|c|c|c|c|c|c|c|c|c|c|c|c|}
\hline
Frame ID & 1 & 2 & 3 & 4 & 5 & 6 & 7 & 8 & 9 & 10 & 11 & 12 & 13 & 14 & 15 & 16 & \textbf{Mean 4} & \textbf{Mean 16}\\
\hline
SRCNN & 97.19 & 92.14 & 95.54 & 94.19 & 86.53 & 100.21 & 103.49 & 94.93 & 93.71 & 95.44 & 98.27 & 97.26 & 109.08 & 68.78 & 76.55 & 106.63 & 94.77 & 94.37\\
\hline
FAST & 97.19 & 0.12 & 0.17 & 0.14 & 0.32 & 0.12 & 0.16 & 0.14 & 0.36 & 0.13 & 0.17 & 0.15 & 0.41 & 0.13 & 0.19 & 0.15 & 24.41 & 6.25\\
\hline
Speed up & 1$\times$ & 776$\times$ & 556$\times$ & 660$\times$ & 268$\times$ & 819$\times$ & 663$\times$ & 691$\times$ & 259$\times$ & 754$\times$ & 595$\times$ & 653$\times$ & 266$\times$ & 518$\times$ & 399$\times$ & 731$\times$ & 3.9$\times$ & 15.1$\times$\\
\hline
\end{tabular}}}
\caption{Run time (in msec) and acceleration of the FAST framework using SRCNN on the test sequence "FourPeople".}
\label{tab:time_four_people}
\end{table*}

\begin{figure}
\centerline{\includegraphics[width=\columnwidth]{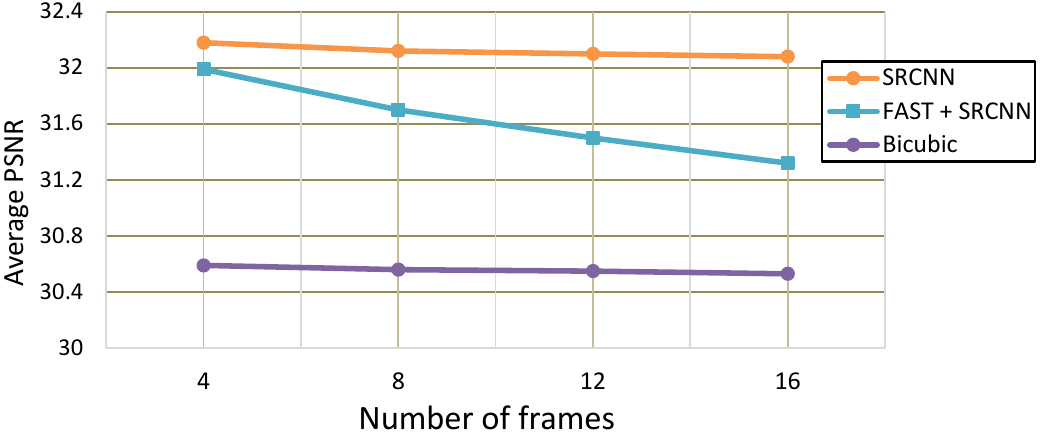}}
\caption{When $QP=22$, the average PSNR across all the test sequences drops as FAST transfers more frames if no thresholding on accumulated error (\ie, adaptive transfer reset) is performed.}
\label{fig:qp_22_degrade}
\end{figure}

The previous results were evaluated on video sequences with a quantization parameter ($QP$) setting of 27; this is a typical quantization setting used in video compression. To evaluate the sensitivity to the accumulated transfer error, we reduce the $QP$ to 22, which reduces the compression ratio resulting in a higher quality video.  Without adaptive transfer reset, we notice a large drop in PSNR as shown in Figure~\ref{fig:qp_22_degrade}, where the $16^{th}$ frame has around 0.7dB lower PSNR than SRCNN, which is much worse than the $0.2$dB for $QP$ of 27. As discussed earlier, this error accumulation is expected when the quantization used during compression is low since there are more high frequency components in the residual that are not preserved by the bicubic interpolation.

\begin{figure}
\centering{	
    \subfigure[Average PSNR vs. transfer ratio]{
		\includegraphics[height=0.33\columnwidth]{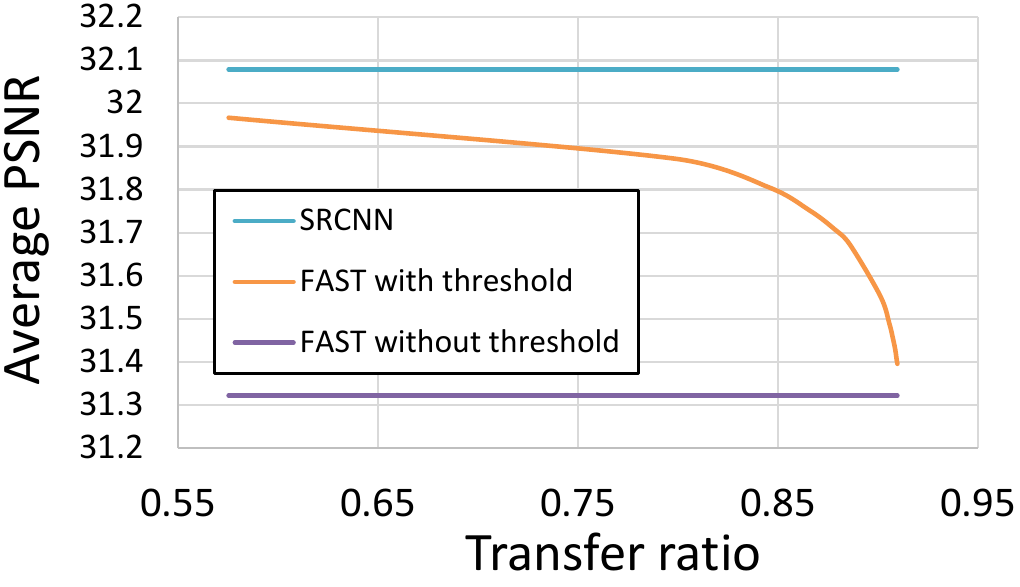}
		\label{fig:psnr_vs_transfer_ratio}
	}
    \subfigure[PSNR drop vs. acceleration ratio]{
		\includegraphics[height=0.33\columnwidth]{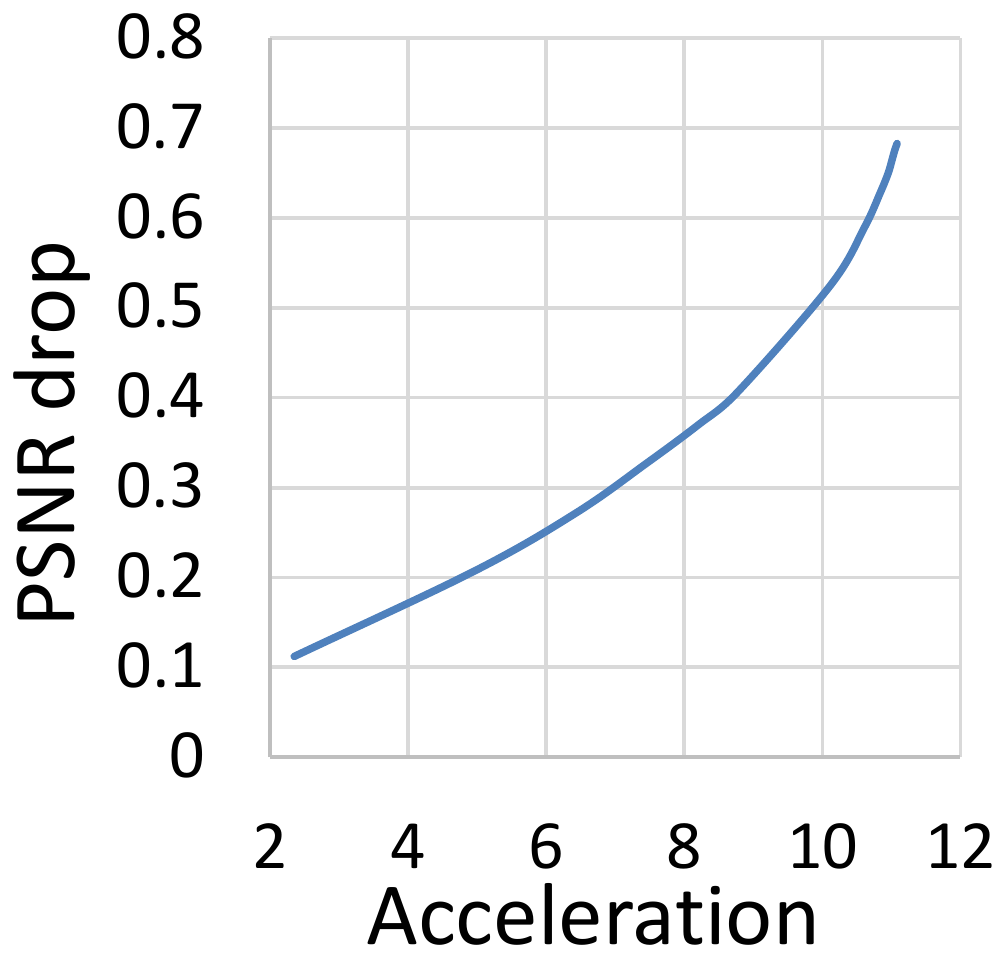}
		\label{fig:psnr_drop_vs_acceleration}
	}

}
\caption{Trade-off between visual quality and speed for FAST when $QP=22$ for a GOP size of 16.}
\label{fig:qp_22_threshold}
\end{figure}

Applying the adaptive transfer reset reduces this PSNR drop at the cost of acceleration as shown in Figure~\ref{fig:qp_22_threshold}.  In these plots, we sweep the threshold discussed in Section~\ref{sec:transfer_reset}. Decreasing the threshold reduces the number of blocks that can be transferred, thus improving the super-resolution visual quality while lowering the acceleration ratio. When more than $80\%$ of the blocks are transferred for $5\times$ acceleration, the PSNR drop is only around $0.2$dB. Note that when the threshold is set to zero, there are still more than $50\%$ blocks transferred, resulting in a $2\times$ acceleration; this is due to the fact that on average, in a typical encoded video, over 50\% of the pixels have zero residual; in all these cases, the blocks are transferred regardless of the threshold.  Accordingly, the minimum transfer ratio plotted in Figure~\ref{fig:qp_22_threshold} is 0.55; the PSNR would remain the same below this ratio.

In summary, FAST accelerates SR algorithms by an order of magnitude in typical video quality settings. Even in the challenging setting with low quantization and the chained GOP structure (IPPP), FAST still achieves a significant ($5\times$) acceleration with an acceptable PSNR loss around $0.2$dB. This validates the effectiveness of FAST on accelerating a SR algorithms across all settings in practice.

\section{Conclusions} 
\label{sec:end}
This paper offers new insight on how to view the input to computer vision and image processing algorithms. Most visual content is stored in a compressed format; thus, rather than viewing the input as only an array of pixels, we should also take into account that it also contains useful information about the structure of the visual content. 

In this paper, we have shown how exploiting this information in the FAST framework can accelerate various SR algorithms by up to 15$\times$ with only $0.2$dB loss. Thus, the FAST framework is an important step towards enabling high visual quality SR algorithm in real-time for ultra-high resolution displays with low computational complexity.

FAST also demonstrates how using non-overlapping block division with deblocking filter reduces computation and avoids artifacts near the block boundary. This approach can potentially be extended to other SR and image processing algorithms. 

As far as we know, FAST is the first technique to use the embedded information in compressed video to accelerate super-resolution on video. We believe that this approach can also be extended to accelerate other computer vision and image processing algorithms, which is becoming increasingly important as the latest state-of-the-art algorithms tend to use CNNs with high computational complexity.

\clearpage
\bibliographystyle{ieeetr}
\bibliography{IEEEabrv,egbib}

\clearpage
\appendix
\noindent
\textbf{Supplementary Material for \textit{FAST: A Framework to Accelerate Super-Resolution Processing on Compressed Videos}}

\section{Formulating Super-Resolution Transfer} 
\label{sec:fast_formulation}

\begin{figure*}
\centerline{\includegraphics[width=0.9\textwidth]{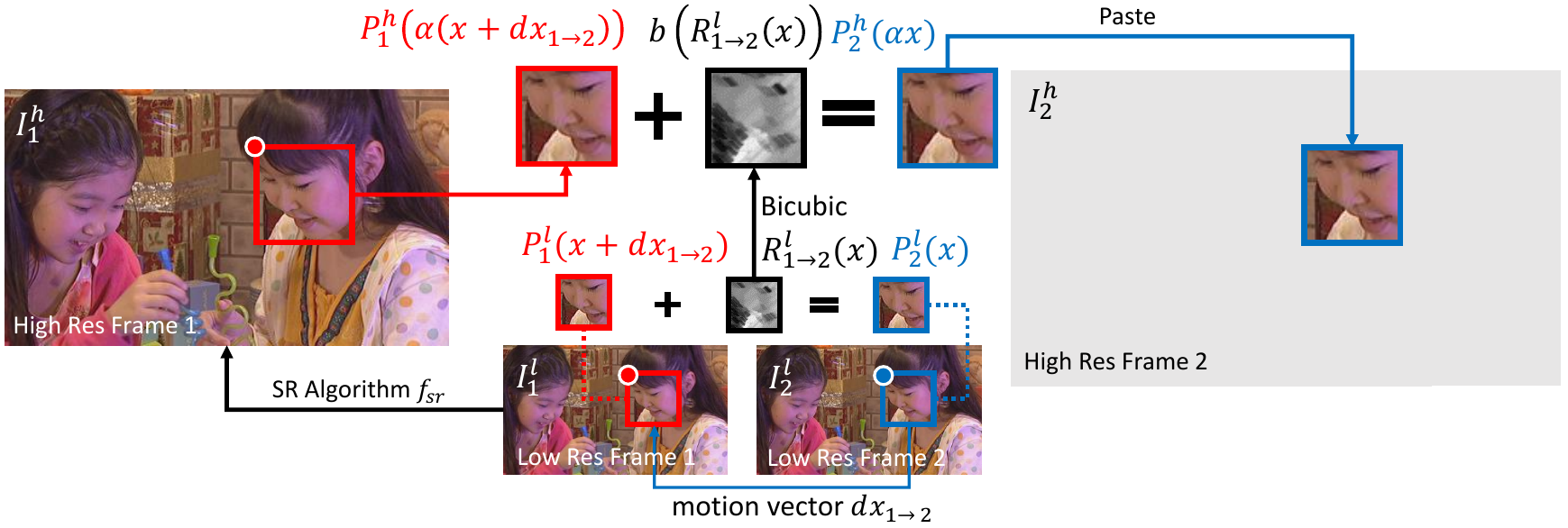}}
\caption{Pipeline of the FAST algorithm.}
\label{fig:FAST_pipeline}
\end{figure*}

This section describes how FAST upsamples a compressed video to be $\alpha\in \mathbb{Z}^+$ times larger. For simplicity, we only consider two adjacent frames, denoted by $I_1^l$ and $I_2^l$ (see Figure~\ref{fig:FAST_pipeline}), where \emph{all} blocks in $I_2^l$ are predicted by motion compensated blocks in $I_1^l$. Here, the subscript indicates the frame index, and the superscript $l$ stands for lower resolution. The goal is to compute the higher resolution images $I_1^h$ and $I_2^h$, where the superscript $h$ stands for higher resolution. We apply a SR algorithm $f_{sr}(\cdot)$ on $I_1^l$ to get $I_1^h = f_{sr}\left(I_1^l\right)$. Instead of applying $f_{sr}(\cdot)$ on the second frame $I_2^l$ to get $I_2^h$, FAST exploits temporal correlation with information from the syntax elements to transfer pixels from $I_1^h$ to $I_2^h$.

FAST repeats the transfer for $I_2^h$ block by block. We now describe how FAST upsamples a low-resolution target block $P_2^l(x)\in\Re^{H\times W}$ in $I_2^l$, where $x\in\Re^2$ is the position within the low resolution frame $I_2^l$, to a higher resolution block $P_2^h(\alpha x)\in\Re^{\alpha H\times \alpha W}$, which is $\alpha$ times larger, where $\alpha x\in\Re^2$ is the position within the high resolution frame $I_2^h$. The decoder tells FAST that $P_2^l(x)$ is predicted by a source block from $I_1^l$ with motion vector $dx_{1\to2}$ with a prediction residual of $R_{1\to2}^l(x)\in\Re^{H\times W}$. From the definition of motion compensation, we have 

\begin{small}
\begin{equation}
\label{eqn:low_mc}
P^l_2(x) = P^l_1(x + dx_{1\to2}) + R_{1\to2}^l(x)
\end{equation}
\end{small}

\noindent Here the subscripts of $dx_{1\to2}$ and $R_{1\to2}$ highlight the prediction direction from frame 1 to frame 2.  Furthermore, let $P_1^h(\alpha x) = f_{sr}\left(P_1^l(x)\right)$. Using Taylor expansion, FAST approximates $P^h_2(\alpha x)$ by

\begin{small}
\begin{equation}
\begin{split}
\label{eqn:mc_taylor}
P^h_2(\alpha x) &= f_{sr}\left(P^l_1(x + dx_{1\to2}) + R_{1\to2}^l(x)\right)\\
&\approx f_{sr}\left(P^l_1(x + dx_{1\to2})\right) + \left<\nabla f_{sr},R_{1\to 2}^l(x)\right> 
\end{split}
\end{equation}
\end{small}

$f_{sr}\left(P^l_1(x + dx_{1\to2})\right)$ is the motion compensated block on $I_1^h$, which is the output of FAST on the first frame.  In~\cite{yang2013fast}, it is claimed that $\left<\nabla f, R_{1\to2}^l(x)\right>$ is close to the upsampling and sharpening operator. In practice, we observe that bicubic upsampling on $R_{1\to2}^l(x)$ is a sufficient approximation of $\left<\nabla f_{sr}, R_{1\to2}^l(x)\right>$, and requires less computation. Let $P_i^{h\prime}(x)$ denote the output of FAST, and let $b(\cdot)$ denote the $\alpha$-times bicubic upsampler. FAST outputs

\begin{small}
\begin{equation}
\label{eqn:transfer}
P^{h\prime}_2(\alpha x) \approx P^{h\prime}_1\left(\alpha(x + dx_{1\to2})\right) + b\left(R_{1\to2}^l(x)\right)
\end{equation}
\end{small}

In summary, \textit{FAST transfers the high-resolution output of $f_{sr}$ on frame 1 to frame 2}. FAST requires: (1) $P^{h\prime}_1\left(\alpha(x + dx_{1\to2})\right)$ obtained using motion compensation on $I_1^h$ with motion vector $\alpha dx_{1\to2}$ at $\alpha x$, and (2) the bicubicly upsampled residual to obtain each block in $I_2^h$. Since $dx_{1\to2}$ may be fractional, $P_1^{h\prime}\left(\alpha(x + dx_{1\to2})\right)$ may also require interpolation. Observe that FAST skips applying $f_{sr}$ to the second frame (and subsequent frames) entirely, and all operations in FAST are similar in complexity to bicubic upsampling. This gives significant savings in computation compared to that of modern SR algorithms such as SRCNN~\cite{dong2014learning}.

\subsection{Accumulated Transfer Error Across Frames}
\label{sec:accumulated_error}
\begin{figure}
\centering{
    \subfigure[Frame 1]{
    \includegraphics[height=0.25\columnwidth]{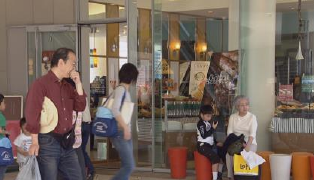}
    } 
    \subfigure[Frame 16]{\includegraphics[height=0.25\columnwidth]{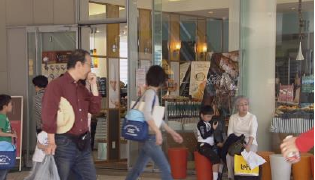}
    }
\\
    \subfigure[PSNR drop visualization]{\includegraphics[height=0.5\columnwidth]{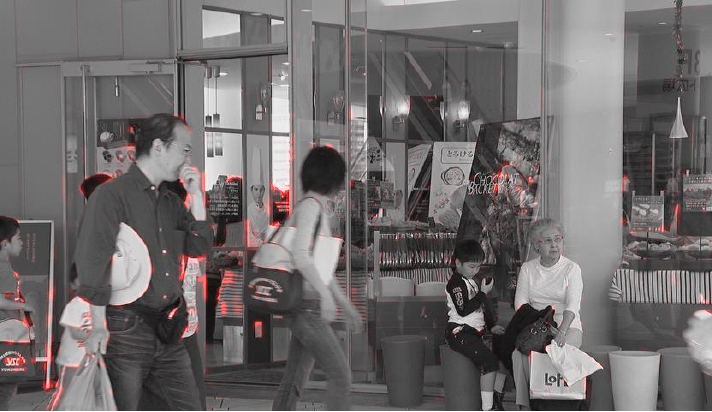}
    }
}
\caption{(a) and (b): the first and the sixteenth frame of the sequence \emph{BQMall} in which the pedestrians are walking in front of a shopping mall. (c): an inbetween frame is dimmed and blended with a layer in red showing the high blockwise PSNR drop of FAST compared with SRCNN. Note that areas with high PSNR drop are mostly the neighborhood of moving occluding edges. Small PSNR drop is observed on the static occluding edges. Non-occluding textures, including the moving faces, are free from degradation.}
\label{fig:qp_22_vis}
\end{figure}

Upsampling the residual to the higher resolution using bicubic interpolation as an approximation introduces errors, as is shown in Eq.~\eqref{eqn:mc_taylor} and~\eqref{eqn:transfer}. In a chained GOP structure (I-P-P-P), a frame is always predicted by the previous frame; thus the error gets accumulated. If large enough, this error can degrade the visual quality of the transferred frame. Such error is visualized in Figure~\ref{fig:qp_22_vis}. To address this, we propose a model to quantify the accumulated error, on which we can perform a threshold to decide whether we should stop the transfer for each block in the video.

Specifically, let us denote the result of FAST at frame $t$ by $P^{h\prime}_t(x)$ as opposed to the SR result $P_t^h(x)$. Under this notation, Eq.~\eqref{eqn:mc_taylor} and~\eqref{eqn:transfer} can be rewritten as
\begin{small}
\begin{equation}
\begin{split}
\label{eqn:FAST_block_def}
P^h_{t}(\alpha x) =& f_{sr}\left(P^l_{t-1}(x + dx_{t - 1\to t})\right) + \left<\nabla f_{sr},R_{t-1\to t}^l(x)\right> \\
P^{h\prime}_{t}(\alpha x) =& P^{h\prime}_{t - 1}\left(\alpha(x + dx_{t - 1\to t})\right) + b\left(R_{t - 1 \to t}^l(x)\right)
\end{split}
\end{equation}
\end{small}

We define the transfer error as $e_t^h(\alpha x) = P^{h\prime}_t(\alpha x) - P_t^h(\alpha x)$. Substituting Eq.~\eqref{eqn:FAST_block_def} into it, we get
\begin{small}
\begin{equation}
\begin{split}
&e_t^h(\alpha x) = P^{h\prime}_t(\alpha x) - P_t^h(\alpha x)\\
=&\left(P^{h\prime}_{t-1}\left(\alpha(x + dx_{t-1\to t})\right) + b\left(R_{t-1\to t}^l(x)\right)\right)-\\
& \left(f_{sr}\left(P^l_{t-1}(x + dx_{t - 1\to t})\right) + \left<\nabla f_{sr},R_{t - 1\to t}^l(x)\right>\right)\\
=&\left(P^{h\prime}_{t-1}\left(\alpha(x + dx_{t-1\to t})\right) - P_{t-1}^h(\alpha(x + dx_{t-1\to t}))\right) + \\
&\left(b\left(R_{t-1\to t}^l(x)\right) - \left<\nabla f_{sr},R_{t - 1\to t}^l(x)\right>\right)\\
=&e_{t - 1}^h(\alpha (x+dx_{t-1\to t})) + \\
&\left(b\left(R_{t-1\to t}^l(x)\right) - \left<\nabla f_{sr},R_{t - 1\to t}^l(x)\right>\right)
\end{split}
\end{equation}
\end{small}

As previously discussed, $\left<\nabla f, R_{t - 1\to t}^l(x)\right>$ is close to an upsampling and sharpening operator; hence, $\left<\nabla f, R_{t - 1\to t}^l(x)\right> - b\left(R_{t-1\to t}^l(x)\right)$ represents the high passed residual, which can be approximated by the Laplacian of the residual $\Delta R_{t - 1\to t}^h(\alpha x)$. Therefore,
\begin{equation}
e_t^h(\alpha x) = e_{t - 1}^h(\alpha (x+dx_{t-1\to t})) - \Delta R_{t - 1\to t}^h(\alpha x)
\end{equation}

To reduce computation, we can express the error in the lower-resolution domain
\begin{equation}
\label{eqn:accumulated_error_model}
e_t^l(x) = e_{t - 1}^l (x+dx_{t-1\to t}) - \Delta R_{t - 1\to t}^l(x)
\end{equation}

This suggests that the transfer error of a new frame is the motion compensated transfer error from the previous error plus the Laplacian of the residual of the current frame. Therefore, the Laplacian of the residual gets accumulated. The computation for this error only involves simple interpolation and addition. For each block, FAST computes the average absolute magnitude of the accumulated error. If it is above a given threshold, the transfer is stopped and the SR algorithm is applied to its lower-resolution decoded block. This resets the accumulated error to zero.

A few observations regarding the accumulated error of different types of blocks:
\begin{itemize}
    \item Smooth blocks do not introduce accumulated error in general.
    \item The accumulated error is not dependent on the low-frequency component of the residual; thus the accumulated error should still be small for blocks with changing brightness.    
    \item A non-occluding edge is often well-predicted from the same edge in the previous frame, so the residual magnitude should be small whether the block is moving or not.
    \item If a block lies on amoving occluding edges in a highly textured area, the error can accumulate rapidly due to the new and unpredictable textures that come out of the occlusion.
\end{itemize}

\end{document}